\documentclass[10pt,twocolumn,letterpaper]{article}

\usepackage{wacv}              

\usepackage{graphicx}
\usepackage{amsmath}
\usepackage{amssymb}
\usepackage{booktabs}

\usepackage{multirow}
\usepackage{graphicx}
\usepackage{float}
\usepackage[normalem]{ulem}
\useunder{\uline}{\ul}{}

\usepackage{enumitem}
\usepackage[accsupp]{axessibility} 

\usepackage[pagebackref,breaklinks,colorlinks]{hyperref}

\usepackage[capitalize]{cleveref}
\crefname{section}{Sec.}{Secs.}
\Crefname{section}{Section}{Sections}
\Crefname{table}{Table}{Tables}
\crefname{table}{Tab.}{Tabs.}

\def\wacvPaperID{1305} 
\def\confName{WACV}
\def\confYear{2025}

\begin{document}

\title{N-Modal Contrastive Losses with Applications to \\ Social Media Data in Trimodal Space}

\author{William Theisen\\
University of Notre Dame\\
{\tt\small wtheisen@nd.edu}
\and
Walter Scheirer\\
University of Notre Dame\\
}
\maketitle

\begin{abstract}

The social media landscape of conflict dynamics has grown increasingly multi-modal. Recent advancements in model architectures such as CLIP have enabled researchers to begin studying the interplay between the modalities of text and images in a shared latent space. However, CLIP models fail to handle situations on social media when modalities present in a post expand above two. Social media dynamics often require understanding the interplay between not only text and images, but video as well. In this paper we explore an extension of the contrastive loss function to allow for any number of modalities, and demonstrate its usefulness in trimodal spaces on social media. By extending CLIP into three dimensions we can further aide understanding social media landscapes where all three modalities are present (an increasingly common situation). We use a newly collected public data set of Telegram posts containing all three modalities to train, and then demonstrate the usefulness of, a trimodal model in two OSINT scenarios: classifying a social media artifact post as either pro-Russian or pro-Ukrainian and identifying which account a given artifact originated from. While trimodal CLIP models have been explored before (though not on social media data), we also display a novel quadmodal CLIP model. This model can learn the interplay between text, image, video, and audio. We demonstrate new state-of-the-art baseline results on retrieval for quadmodel models moving forward.

\end{abstract}

\begin{figure}[!t]
\centering
\includegraphics[width=0.90\linewidth]{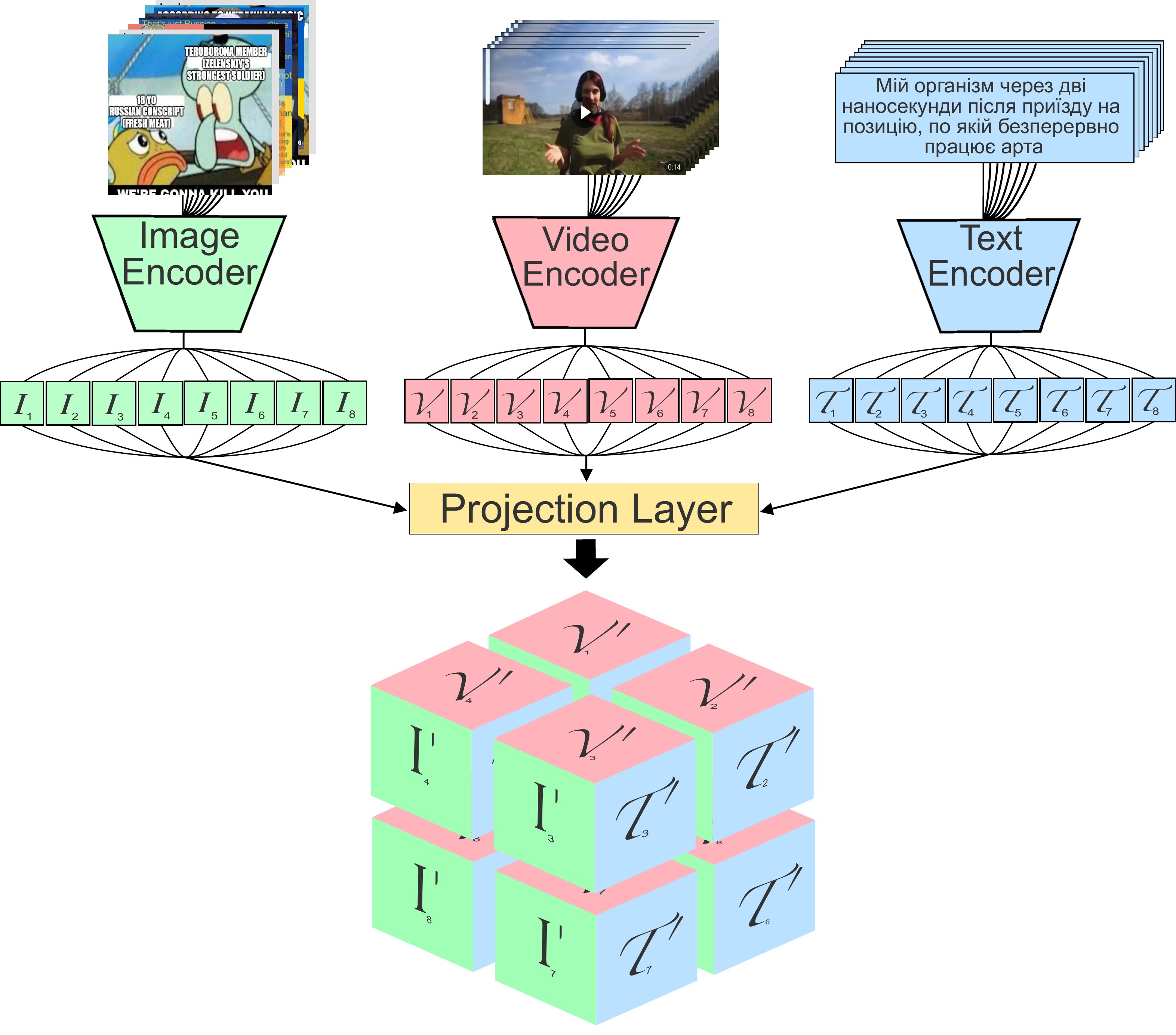} 
\caption{An intuitive visualization of contrastive loss expanded to a trimodal space, with the optimization happening across a cube of similarities rather than a 2-dimensional grid in order to account for the multi-modal properties of a social media post containing not only images and text, but video as well. After training a shared projection layer embeddings from all modalities are projected into a shared latent space, with artifacts from the same post being close to each other.}
\vspace{-5mm}
\label{fig:tease}
\end{figure}

\section{Introduction}
\label{sec:intro}

As social media continues to grow so to has open-source Intelligence (OSINT) played an increasingly large role in government's responses to international conflicts. The recent invasion of the Ukraine by Russia has been accompanied by a truly digital form of warfare, with social media literally leading to "boots-on-the-ground" action. An example of this is the Ukrainian government using a social media post by a Russian soldier to target a missile strike on his platoons location~\cite{The_Economist_2023}. Unfortunately most OSINT work requires the majority of the heavy lifting to be done manually by human operators. Understanding not only the inter-post context of a social media artifact, let alone the surrounding context, is a difficult problem. One such issue preventing more advanced computational success in this scenario is the difficulty of understanding multi-modal posts on social media. Social media posts are no longer just small snippets of text, and indeed posts containing images and videos have surpassed the amount of text that is now posted on social media [*]. To further computational OSINT capabilities multi-modal understanding is critical.

One recent advance in multi-modal understanding in the computer sciences has been the introduction of Contrastive Language-Image Pre-training (hereafter CLIP)~\cite{radford2021learning}. By training models on pairs of images and text the model can learning the similarities between the two modalities. CLIP models have become ubiquitous in the field and form the foundation of many exciting new multimodal AI tools today such as Stable Diffusion~\cite{stablediffusion} and DALLE~\cite{dalle}. These models work extremely well for relating images and text together. Unfortunately, as discovered by Theisen and Scheirer~\cite{theisen2023cclip}, CLIP models trained on non-social media data do not work when applied to social media data. This means that models for use on social media data need to be trained from scratch. While CLIP models for three modalities have been explored (lightly) in the literature, they've yet to be applied to social media. Additionally we formalize the extension of CLIP loss to dimensions higher than three and give a novel demonstration of what a quadmodal model would look like.
 
 In short, the paper makes the following contributions:
\begin{enumerate}[noitemsep,nolistsep]
    \item An formal extension of two contrastive losses to 3+ modalities
    \item A new data set of social media posts containing videos and text, out of which can be synthesized a trimodal data set
    \item Publicly available trimodal models
    \item A demonstration of trimodal CLIP applications applied to social media, via a binary stance classifier and a multiclass account provenance classifier
    \item A novel proof-of-concept quadmodal CLIP model, providing a new baseline for quadmodal models in the future
\end{enumerate}

\section{Related Work}
\label{sec:rel_wrk}

\textbf{Motivating Works:} There are a number of pipelines intending to aid social media understanding that have been published. One part of computation social media analysis is understanding the interplay between different types of items posted on social media such as images, videos, and text. Early work in social media image understanding was done by Zannettou et al.~\cite{zanne}, Beskow et al.~\cite{beskow}, Dubey et al.~\cite{dubey}, and Theisen et al.~\cite{theisen}. All of these were published prior to CLIP, and thus suffer from focusing on only a single modality (images). A similar work that expands across modalities is "Few-shot Learning for Multi-modal Social Media Event Filtering" by Nascimento et al.~\cite{nascimento2022fewshot}. This work uses CLIP to embed image-text pairs into the same space to increase the ability to detect "events" from social media posts. Another such project is called "MEWS: Misinformation Early Warning System", demo'ed by Ford et al~\cite{ford2022mews}. While this project extracts embeddings for more than two types of modalities, it keeps them in separate latent spaces and instead does the comparison post-hoc.

\textbf{Autoencoders:} The models used in this paper are merely built on top of pre-existing work. We use four different off the shelf models for the four modalities studied in this work: video, image, text, and audio. For vision we use a masked autoencoder from Meta~\cite{DBLP:journals/corr/abs-2111-06377}. We pair with this a multilingual distilBert model~\cite{DBLP:journals/corr/abs-1910-01108}~\cite{reimers2020making}. For video we use another masked autoencoder: videoMAE~\cite{tong2022videomae}. Audio feature extraction was done using wav2vec2~\cite{DBLP:journals/corr/abs-2006-11477}. Having strong features is the foundation on top of which contrastive learning operates. We did not task pre-train these autoencoders in this work.

\textbf{Multimodal Embedding Models:}
Early work in multimodal understanding was focused primarily on "fusion" techniques, or methods of combining relatively unrelated features post-hoc~\cite{PORIA201798}~\cite{5871582}. The two most common fusion techniques were early fusion and late fusion. Early fusion ~\cite{5871582}~\cite{poria-etal-2017-context} relies on concatenating feature vectors, as can be seen in~\cite{theisen}. Late fusion uses a weighted average technique to consider the monomodal feature vectors differently~\cite{mosi}~\cite{5674019}~\cite{Zhao_2023}. While fusion worked, it was rather inelegant and struggled to capture the interplay between modalities at a fundamental level. As fusion techniques continued to be developed, tensor-based fusion ~\cite{mai-etal-2019-divide}~\cite{DBLP:journals/corr/ZadehCPCM17}~\cite{NEURIPS2019_f56d8183} and low-rank modality fusion rose to the fore~\cite{DBLP:journals/corr/abs-1806-00064}. These methods used an outer product method to learn both intra- and inter-modality features. As contrastive losses have become more prevalent, fusion techniques have fallen by the wayside. 

Prior to the introduction of contrastive loss, triplet loss had been shown to work on bimodal data sets with slight modifications. In 2018 Deng et al.~\cite{8331146} showed that triplet loss could be used to facilitate hashing methods for cross modal retrieval. Wang et al.~\cite{DBLP:journals/corr/WangLL17} demonstrated that expanding the triplet loss with additional terms and improving neighborhood contstraints could increase the retrieval results across modalities.

This work is heavily based on prior advances in multimodal machine learning. First and foremost is the introduction of Contrastive Image-Language Pre-training (CLIP) by Radford et al. in 2021~\cite{radford2021learning}. CLIP allows for multimodal pairs (original image-text) pairs to be contrasted into a shared latent space. The algorithm introduced in CLIP led to a flurry of papers focused on multimodal understanding such as VideoCLIP~\cite{DBLP:journals/corr/abs-2109-14084}, Wav2CLIP~\cite{wav2clip} (a method for learning audio representations via CLIP), Clip-nerf~\cite{nerfclip} (allowing for the manipulation of neural radiance fields), and C-CLIP~\cite{theisen2023cclip} (a version of CLIP trained explicitly for social media). C-CLIP is the most similar work, and the one we compare against, due to its focus on social media.

Extending contrastive loss to three modalities has been previously explored in the literature~\cite{vatt}~\cite{tupleinfonce}~\cite{multimodalversatile}. Mai et al.\cite{hybridcontrastive} proposed what they termed a "Hybrid Contrastive" model for projecting audio, images, and text into a shared latent space with the goal of using the combined vectors to perform sentiment analysis. Another exploration of trimodal learning via contrastive loss was by Ruan et al.~\cite{ruan2024tricolo} in 2024. In their work "TriCoLo: Trimodal Contrastive Loss for Text to Shape Retrieval" they showed that one can successfully map three modalities: voxel shapes of tables and chairs, images of these voxel shapes from different angles, and text descriptions of the shapes into the same latent space. Our experimental setup is similar to theirs, but we instead optimize for social media data, due to the task discrepancy outlined in Theisen and Scheirer~\cite{theisen2023cclip}.

\textbf{Classifiers — Binary and Multiclass:} To demonstrate the usefulness of the models we demonstrate two classifiers, a binary classifier and a multiclass classifier. Three common techniques were tested using the features extracted from our models, for both binary and multiclass applications: Naive Bayes~\cite{bayes}, Random Forests~\cite{forests}, and SVMs~\cite{svm} (Support Vector Machines). In addition to these three, we train two basic models, one binary and one multiclass following the outline provided by Tam~\cite{Tam_2023_binary}~\cite{Tam_2023_multi}.

\section{N-modal Losses}
\label{sec:theory}
 
 For this work we explore the extension of two common loss functions used in bimodal learning, triplet loss and contrastive loss, in a manner similar to Ruan et al.~\cite{ruan2024tricolo}. Given below is a brief overview of the two loss functions at the bimodal level, followed by an explanation of our extension of these two losses to a higher number of modalities.

\textbf{Bimodal Triplet Loss:}
Triplet loss involves three elements: an anchor, a positive example, and a negative example. The goal is to ensure that the anchor is closer to the positive example than to the negative example in the embedding space. Triplet loss can be mathematically expressed as:

\begin{align}   
& L_{triplet} = \max\{sim(A, P) \nonumber \\
& - sim(A, N) + margin, 0\}
\end{align}

where \( A \) represents the anchor, \( P \) the positive example, \( N \) the negative example, \( sim() \) is a similarity function, and \( margin \) is a hyperparameter. When used for bimodal learning, the positive sample and the anchor are from the two different modalities associated with the ground-truth pair, while the positive sample and the negative sample are from the same modality but different pairs. In this manner, the loss function ensures the cross-modal positive samples are closer than the cross-modal negative samples, therefore allowing for bimodal comparisons in the latent space.

\textbf{N-Modal Triplet Loss:}
A basic extension of triplet loss to include additional modalities beyond two is relatively straight forward. One can simply iterate over the different modality combinations to generate comparisons across all modality pairings. We can generalize an N-modal triplet loss function over a batch of size \(N\) and \(M\) modalities as:

\begin{equation}
L_{total} = \alpha \sum_{i=0}^{N-1} \sum_{a=1}^{M} \sum_{p=1}^{M} \sum_{q=1}^{M} L(e^{a}_i, e^{p}_i, e^{q}_{(i+1) \mod N})
\end{equation}

where:
\begin{itemize}
    \item \(e^{a}_i, e^{p}_i\) are the embeddings for the anchor and positive examples in the \(i\)-th element, selected from the set of \(M\) modalities,
    \item \(e^{q}_{(i+1) \mod N}\) is the embedding for the negative example from the cyclically next element in the batch, ensuring a diverse selection,
    \item \(L(a, p, n)\) represents the triplet loss between an anchor \(a\), a positive \(p\), and a negative \(n\) example,
    \item \(\alpha\) is a scaling factor applied to the total loss.
\end{itemize}

The indices \(a, p, q\) iterate over the modalities (1 through \(M\)), allowing for all combinations of anchor, positive, and negative selections within and across modalities. So much like in bimodal triplet loss, the positive, anchor, and negative are computed across modalities. When three or more modalities are present, one can simple brute-force all triplets across modalities for which the triplet loss could be computed. Unsurprisingly this is a costly loss to compute, as with three modalities it requires 9 comparisons, and with four it would require 16. So while it works, it scales very poorly, requiring N\textsuperscript{2} computations, where N is the number of modalities. This appears directly in the runtimes of training a model using this loss and can be seen in Table~\ref{tab:runtimes}.

\textbf{Bimodal CLIP Loss:}
Constrasted to triplet loss, CLIP loss is a method used to learn embeddings by contrasting positive pairs (similar or related items) against all other negative pairs (dissimilar or unrelated items). The goal is to ensure that positive pairs are closer in the shared embedding space than negative pairs.

The loss is defined as:

\begin{align}
    L_{CLIP} = -\frac{1}{2n} \sum_{i=1}^n \Biggl[\log \frac{e^{sim(I_i, T_i)/\tau}}{\sum_{j=1}^n e^{sim(I_i, T_j)/\tau}} \nonumber \\ 
    + \log \frac{e^{sim(T_i, I_i)/\tau}}{\sum_{j=1}^n e^{sim(T_j, I_i)/\tau}}\Biggr]    
\end{align}

where \( I_i \) and \( T_i \) are the embeddings of the image and text in the \( i^{th} \) pair, respectively~\cite{radford2021learning}~\cite{DBLP:journals/corr/abs-2010-00747}.

\textbf{N-Modal CLIP Loss:} N-Modal CLIP Loss extends the traditional concept of contrastive loss to accommodate more than two modalities. In N-Dimensional CLIP Loss, data from N different modalities are projected into a shared embedding space. Similarly to the bimodal loss shown above, the objective is to minimize the distance between corresponding tuples from different modalities while maximizing the distance between non-corresponding tuples. We take as our "ground-truth" tuples the N artifacts present in a single social media post, where each artifact is from a different modality. If bimodal contrastive loss can be visualized as a grid of pairs, trimodal contrastive loss could be visualized as a cube (refer to Figure~\ref{fig:tease}). More than three modalities begins to be harder to visualize, but the process of the extensions is the same as going from 2 to 3 modalities.

The loss function for N-Dimensional CLIP Loss is given by:
\begin{equation}
    \begin{aligned}
        & L_{N-contrastive} = \\
        & \frac{1}{N}\sum_{i=1}^N \sum_{j=1, j\neq i}^N -\log \frac{e^{sim(M_i, M_j)/\tau}}{\sum_{k=1, k\neq i}^N e^{sim(M_i, M_k)/\tau}}    
    \end{aligned}
\label{eq:nclip}
\end{equation}

where \( M_i \) represents the embedding from the \( i^{th} \) modality, \( sim() \) is a similarity function (like cosine similarity), and \( \tau \) is a temperature parameter that scales the similarity scores. As per Ruan et al.~\cite{ruan2024tricolo}, the final step of the loss function is summing the pairwise losses. Whereas they pin it to three dimensions we choose to represent this as a summation across any number of dimensions and then averaged by 2N. This is what formally allows our loss to extend to any number of modalities.

\begin{figure*}[t]
\centering
\includegraphics[height=6cm]{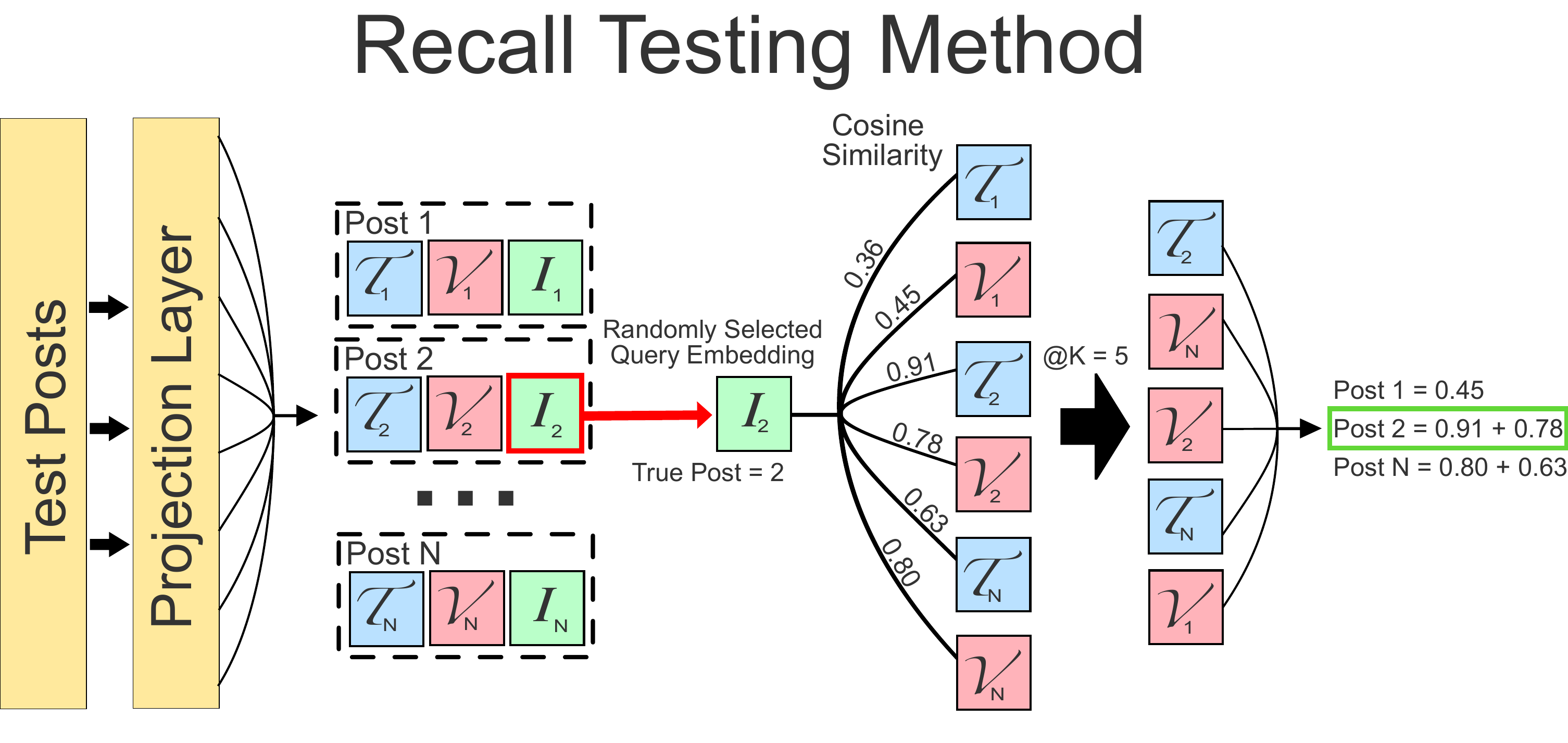} 
\caption{The evaluation method for measuring the recall of our models. The chosen embedding is compared only to those embeddings that are not of the same modality, to highlight the cross-modal abilities of the model. Similarites for the top-K embeddings are then summed together when they are from the same post.}
\vspace{-2mm}
\label{fig:stance_class}
\end{figure*}

\textbf{New Dataset:} Social media datasets with more than two modalities are shockingly difficult to come by. To foster the work we introduce a new dataset collected from Telegram. This dataset consists of Telegram posts containing a video of at least 64 frames and text of at least 5 words. Extending the collection methodology outlined in Theisen and Scheirer~\cite{theisen2023cclip} we collect videos and text instead of images and text. Over the course of the project we collected 69,831 video-text pairs from 33 accounts randomly selected from the list they provided. The discerning reading will notice that this dataset still only contains two modalities, videos and text. A standard video transformer only operates on a small number of frames from the overall video (in this work, 16 frames). This means that we can treat a frame not considered by the video transformer as a separate but related image. While imperfect, it allows us to generate a trimodal dataset out of a bimodal video-text dataset. This new dataset will be made available upon publication, along with any code involved with the work, and all models trained.

\section{Building a Trimodal Models}
\label{sec:method}

To show that the losses can successfully be extended to more than two dimensions, we build two different models based on the N-dimensional triplet and contrastive losses, hereafter referred to as triTRIP and triCLIP respectively, with the amount of training data given as a suffix after the model name.

A bimodal model is typically built with two encoders, one for each modality, on top of which one or more projection layers are placed. It is these projection layers that are then trained using the loss to minimize the distance between the paired encodings. By using the trimodal contrastive loss extension we can perform a similar process using tuples of three items which are encoded separately. For this paper we use the following three encoders: text — 'distilbert-base-multilingual-cased'~\cite{distilbert}, image — 'facebook/vit-mae-base'~\cite{vitmae}, video — 'MCG-NJU/videomae-base'~\cite{videomae}. All three of these encoders are used as-is and are publicly available from HuggingFace. These encoders also have the property of being transformers, meaning their default output length is 768. This means that in theory the projection layer could output upwards of a length 768 vector. Prior to encoding, videos and images were compressed to the size of 244 x 244 pixels, standard practice. Of particular importance was the selection of text encoder. The corpus is multilingual and it was important to select an encoder that could handle the various languages present in the collected data. As recommended by Theisen and Scheirer~\cite{theisen2023cclip}, we use a multilingual distilbert model, in order to handle the multilingual dataset without first needing to translate the texts.

The implementation is based on Shariatnia's implementation of CLIP, but extended to allow for more than two encoders. For the triTRIP and triCLIP implementations, each of the three encoders is given its own projection head, through which we train a projection layer. The projection heads are how the model goes takes a high dimensional vector (in the case of our three encoders, 768 dimensional) and projects it into a shared, lower dimensional, latent space (256 dimensions in our standard implementation of triTRIP and triCLIP). So to train, first the encodings are calculated for every item in the batch, all of these embeddings are then passed through the projection heads assigned to their modality. With the output of the projection heads the loss is then calculated. The runtimes of training are given in Table~\ref{tab:runtimes} for the models at various amounts of training data and epochs. The models were trained using the standard Adam LR optimizer and on a single GPU (out of a heterogeneous collection of). A batch size of 128 was used.

\begin{table}[]
\centering
\resizebox{\columnwidth}{!}{%
\begin{tabular}{ccccc}
\multicolumn{5}{c}{Average Training Time for Models by Epochs (5 Trials)} \\ \hline
\multicolumn{1}{c|}{Model} & \multicolumn{1}{c|}{1} & \multicolumn{1}{c|}{10} & \multicolumn{1}{c|}{50} & 100 \\ \hline
\multicolumn{1}{c|}{triTRIP-100} & \multicolumn{1}{c|}{00:00:14} & \multicolumn{1}{c|}{00:00:31} & \multicolumn{1}{c|}{00:01:36} & \multicolumn{1}{l}{00:01:57} \\
\multicolumn{1}{c|}{triCLIP-100} & \multicolumn{1}{c|}{00:00:12} & \multicolumn{1}{c|}{00:00:13} & \multicolumn{1}{c|}{00:00:21} &  00:00:17\\
\multicolumn{1}{c|}{triTRIP-1k} & \multicolumn{1}{c|}{00:00:22} & \multicolumn{1}{c|}{00:02:24} & \multicolumn{1}{c|}{00:07:16} & 00:23:32 \\
\multicolumn{1}{c|}{triCLIP-1k} & \multicolumn{1}{c|}{00:00:13} & \multicolumn{1}{c|}{00:00:11} & \multicolumn{1}{c|}{00:00:16} & 00:00:19\\
\multicolumn{1}{c|}{triTRIP-10k} & \multicolumn{1}{c|}{00:01:48} & \multicolumn{1}{c|}{00:17:28} & \multicolumn{1}{c|}{01:16:42} & 03:47:24 \\
\multicolumn{1}{c|}{triCLIP-10k} & \multicolumn{1}{c|}{00:00:10} & \multicolumn{1}{c|}{00:00:16} & \multicolumn{1}{c|}{00:00:38} &  00:01:13\\
\multicolumn{1}{c|}{triTRIP-50k} & \multicolumn{1}{c|}{00:11:13} & \multicolumn{1}{c|}{01:53:11} & \multicolumn{1}{c|}{05:59:07} & 18:52:17 \\
\multicolumn{1}{c|}{triCLIP-50k} & \multicolumn{1}{c|}{00:00:15} & \multicolumn{1}{c|}{00:00:41} & \multicolumn{1}{c|}{00:02:33} &  00:05:15\\
\end{tabular}%
}
\caption{The average training times for the trimodal models in HH:MM:SS format. Features were pre-extracted, leading to much shorter training times than one might expect. The triTRIP models take much longer to train due to the modality-pairwise computations required.}
\label{tab:runtimes}
\end{table}

\begin{table*}[t]
\centering
\resizebox{\textwidth}{!}{%
\begin{tabular}{ccccccccccccccccc}
\multicolumn{17}{c}{Average Recall for Models @K by Epochs Trained (5 Trials)} \\ \hline
\multicolumn{1}{c|}{Epochs} & \multicolumn{4}{c|}{1} & \multicolumn{4}{c|}{10} & \multicolumn{4}{c|}{50} & \multicolumn{4}{c}{100} \\ \hline
\multicolumn{1}{c|}{@K} & @1 & @5 & @10 & \multicolumn{1}{c|}{@25} & @1 & @5 & @10 & \multicolumn{1}{c|}{@25} & @1 & @5 & @10 & \multicolumn{1}{c|}{@25} & @1 & @5 & @10 & @25 \\ \hline
\multicolumn{1}{c|}{triTRIP-100} & 1.00\% & 5.00\% & 9.33\% & \multicolumn{1}{c|}{25.80\%} & 1.13\% & 5.73\% & 12.20\% & \multicolumn{1}{c|}{29.27\%} & 2.27\% & 7.47\% & 15.47\% & \multicolumn{1}{c|}{34.67\%} & 3.53\% & 14.27\% & 23.80\% & 46.93\% \\
\multicolumn{1}{c|}{triCLIP-100} & 0.60\% & 4.80\% & 9.60\% & \multicolumn{1}{c|}{24.73\%} & 0.80\% & 6.13\% & 13.13\% & \multicolumn{1}{c|}{30.20\%} & 2.27\% & 11.80\% & 21.20\% & \multicolumn{1}{c|}{42.73\%} & 4.53\% & 16.07\% & 26.60\% & 52.73\% \\
\multicolumn{1}{c|}{triTRIP-1k} & 1.73\% & 6.00\% & 11.40\% & \multicolumn{1}{c|}{30.07\%} & 2.93\% & 12.07\% & 21.40\% & \multicolumn{1}{c|}{41.40\%} & 10.80\% & 34.27\% & 46.07\% & \multicolumn{1}{c|}{68.80\%} & 15.67\% & 41.00\% & 56.33\% & 76.67\% \\
\multicolumn{1}{c|}{triCLIP-1k} & 1.40\% & 7.13\% & 13.60\% & \multicolumn{1}{c|}{29.47\%} & 4.40\% & 18.27\% & 28.60\% & \multicolumn{1}{c|}{49.13\%} & 15.00\% & 40.13\% & 53.27\% & \multicolumn{1}{c|}{72.33\%} & 15.47\% & 40.20\% & 56.53\% & 74.07\% \\
\multicolumn{1}{c|}{triTRIP-10k} & 2.67\% & 11.67\% & 21.00\% & \multicolumn{1}{c|}{44.33\%} & 21.27\% & 45.33\% & 57.47\% & \multicolumn{1}{c|}{75.73\%} & 34.33\% & 62.80\% & 71.40\% & \multicolumn{1}{c|}{86.53\%} & 32.87\% & 60.00\% & 70.13\% & 84.73\% \\
\multicolumn{1}{c|}{triCLIP-10k} & 4.33\% & 17.53\% & 27.13\% & \multicolumn{1}{c|}{49.40\%} & 24.53\% & 55.00\% & 68.60\% & \multicolumn{1}{c|}{83.13\%} & 60.47\% & 76.27\% & 80.87\% & \multicolumn{1}{c|}{88.33\%} & 59.47\% & 73.40\% & 79.07\% & 87.27\% \\
\multicolumn{1}{c|}{triTRIP-50k} & 12.80\% & 36.13\% & 47.40\% & \multicolumn{1}{c|}{68.73\%} & 28.67\% & 61.33\% & 73.13\% & \multicolumn{1}{c|}{86.33\%} & 46.60\% & 73.60\% & 83.33\% & \multicolumn{1}{c|}{{\ul 93.67\%}} & 45.40\% & 73.87\% & 81.00\% & 91.27\% \\
\multicolumn{1}{c|}{triCLIP-50k} & 14.27\% & 38.73\% & 52.20\% & \multicolumn{1}{c|}{71.47\%} & 57.73\% & 73.33\% & 78.07\% & \multicolumn{1}{c|}{86.20\%} & 66.13\% & {\ul 80.33\%} & 85.07\% & \multicolumn{1}{c|}{92.20\%} & {\ul 67.47\%} & 79.60\% & {\ul 85.20\%} & 91.40\%
\end{tabular}%
}
\caption{The recall results of the models trained, averaged across 5 training trials. Evaluated given a single-modality artifact (text, image, or video), the goal was to return the post the artifact originated from. For each of the 5 trials, 300 artifacts were queried. Unsurprisingly the models achieve higher recall when trained on more data for longer, with triCLIP-50k achieving the best results (split across 50 epochs and 100 epochs).}
\label{tab:recall}
\end{table*}

Evaluation of the models was done on a post-level retrieval task. So given a single-modality artifact from a post, the goal is to determine which post it belongs to. Retrieval was measured at K=1, 5, 10, and 25. As the similarities were calculated at the artifact level, they had to be collated to the post level. This was done by simply summing the similarities for each post's artifact if it was returned. A visualization of this can be seen in Fig.~\ref{fig:post_retr}. For each trial there was a population of 100 posts, all of whose artifacts were tested as the query embedding, leading to 300 queries being performed during test time. All tests were then repeated 5 times, including training, to get an average recall score over the 5 trials (deviations of recall can be found in the supplemental material).

\section{Experiments and Results}
\label{sec:exps_res}

In total, 41 models were trained. triTRIP and triCLIP models were trained on training sets of sizes 100, 1000, 10000, and 50000 posts (each of which has three artifacts). Each of these training sizes was trained across a series of 4 epochs: 1, 10, 50, and 100. The runtimes for training can be seen in Table~\ref{tab:runtimes}. The cost of training a triTRIP model is significantly higher than training a triCLIP model due to the modality-pairwise triplet loss computations that need to be performed (the author is willing to allow their optimization skills to be called into question at this time). The features used during training were pre-extracted, so if one were to run the pipeline from scratch, time would need to be added for feature extraction (for reference, extracting features from 50,000 posts took roughly 21 hours). Regrettably, the increased training time required for triTRIP models does not correlate to higher accuracy on the retrieval task, especially at K=1. 

To test the success of the training we evaluated models on a simple retrieval task: given a monomodal social media artifact, we attempt to retrieve the post the artifact belongs to. The model is attempting to maximize the similarity between a tuple of three artifacts. If we are querying a video, then the post the model should return is the one containing said video. However if we compare this video against all three artifacts for every post, the task becomes too easy, as the most similar object would of course be the video itself and would therefore overly bias the recall metric. For this purpose when we query an artifact we query it only across modalities. I.E. if a text is queried it is only compared to images and videos. This means the most similar posts returned are most similar to the text by virtue of their images and videos and avoids achieving a 1.0 recall because the system simply retrieves the post that contains the query data directly. This also highlights the model's ability to learn cross-modal similarities; the goal of the work.

As mentioned in Sec.~\ref{sec:method}, the projection layer could theoretically keep the output vector at 768 dimensions, as this is what all three input vectors are. A natural instinct is to assume that a higher output dimensionality would lead to higher recalls, however per Table~\ref{tab:proj_recall} this appears not to be the case. Our hypothesis is, due to the original task-mismatch of the three transformers used, the higher dimension projection layers leads to the retaining of information extraneous to the task at hand (relating the three modalities to each other). By reducing the output dimensionality the model is able to shed the information that is unrelated to the task, and focus the output embeddings on cross-modality relations. While the original CLIP paper pins the output dimensions at 256, the findings beg the question of "would reducing the output dimensions further increase recall". Therefore in addition to testing recall with output dimensions of 512 and 768, we test recall at 128 and 64. As can be seen in Table~\ref{tab:proj_recall}, a projection layer with size 256 leads to the best results on the majority of K values (and is within the standard deviation of the best result when it is not). Thus, for all other model results reported in this paper, we use the default CLIP projection layer size of 256 dimensions.

In order to ground our results to prior work, we compare with the most similar bimodal CLIP application. Theisen and Scheirer developed a C-CLIP model on a similar dataset of Telegram image-text pairs. Table~\ref{tab:cclip_comp} shows that when trained on the same number of items (image-text pairs in their case, image-text-video tuples in ours) we achieve recall results that are better than other state-of-the-art social media CLIP models, thus demonstrating that the extension of a model to three modalities using the aforementioned loss functions works. An intuitive explanation for the increase in accuracy could be that due to the correct post having two artifacts floating around in the population, it is more likely that given the combination of the two similarity scores for them, the correct post is more likely to be identified. To demonstrate this, we also test our models on bimodal data (image-text pairs), after training on trimodal data. The results of this seem to support the hypothesis, with the image-text only retrieval results being similar to the baseline (though slightly lower).

\begin{table}[]
\centering
\resizebox{\columnwidth}{!}{%
\begin{tabular}{cclllclll}
\multicolumn{9}{c}{Comparison of Recall @K with a Post Population of 100} \\ \hline
\multicolumn{1}{c|}{\multirow{2}{*}{Model}} & \multicolumn{4}{c}{Recall} \\ \cline{2-5}
\multicolumn{1}{c|}{} & @1 & \multicolumn{1}{c}{@5} & \multicolumn{1}{c}{@10} & @25 \\ \hline
\multicolumn{1}{c|}{M-CLIP~\cite{carlsson-EtAl:2022:LREC}} & 23.40\% & 38.80\% & 47.70\% & 66.70\% \\
\multicolumn{1}{c|}{C-CLIP~\cite{theisen2023cclip}} & 16.09\% & 34.89\% & 48.59\% & 68.10\% \\
\multicolumn{1}{c|}{bi-triTRIP-10k} & 12.40\% & 34.60\% & 50.27\% & 73.87\% \\
\multicolumn{1}{c|}{bi-triCLIP-10k} & 10.93\% & 27.20\% & 39.87\% & 66.53\% \\ \hline\hline
\multicolumn{1}{c|}{triTRIP-10k} & 34.33\% & 62.80\% & 71.40\% & 86.53\% \\
\multicolumn{1}{c|}{triCLIP-10k} & 60.47\% & 76.27\% & 80.87\% & 88.33\% \\
\end{tabular}%
}
\caption{A comparison to baseline M-CLIP~\cite{carlsson-EtAl:2022:LREC} and C-CLIP~\cite{theisen2023cclip} models from the literature. The triTRIP and triCLIP models were trained on image-video-text tuples taken from similar Telegram accounts. All models were trained on 10,000 items for 50 epochs, other than M-CLIP (which was trained on 7M image-text pairs). The bi-tri models are evaluated on image-text pairs only, to allow for a more direct comparison to the baseline.}
\label{tab:cclip_comp}
\end{table}

\begin{figure*}[t]
\centering
\includegraphics[width=1.0\linewidth]{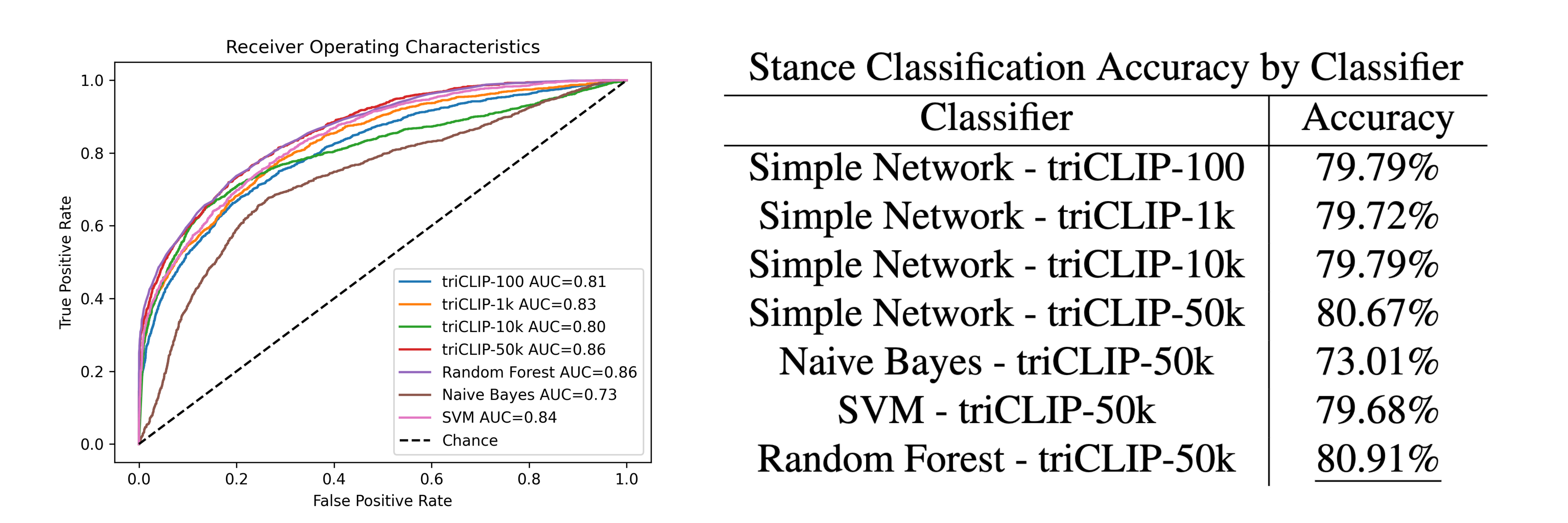} 
\caption{The reciever-operator characteristics curve (ROC) for the stance classifiers on 10,000 posts, along with the area under curve for each classifier. As can be seen, all classifiers achieve results well above the baseline. The accuracy table on the right shows that Random Forests achieve the highest stance classification accuracy at 80.91\%, though all methods other than Naive Bayes were within 1\% of each other.}
\vspace{-2mm}
\label{fig:stance_class}
\end{figure*}

\begin{figure*}[t]
\centering
\includegraphics[width=1.0\linewidth]{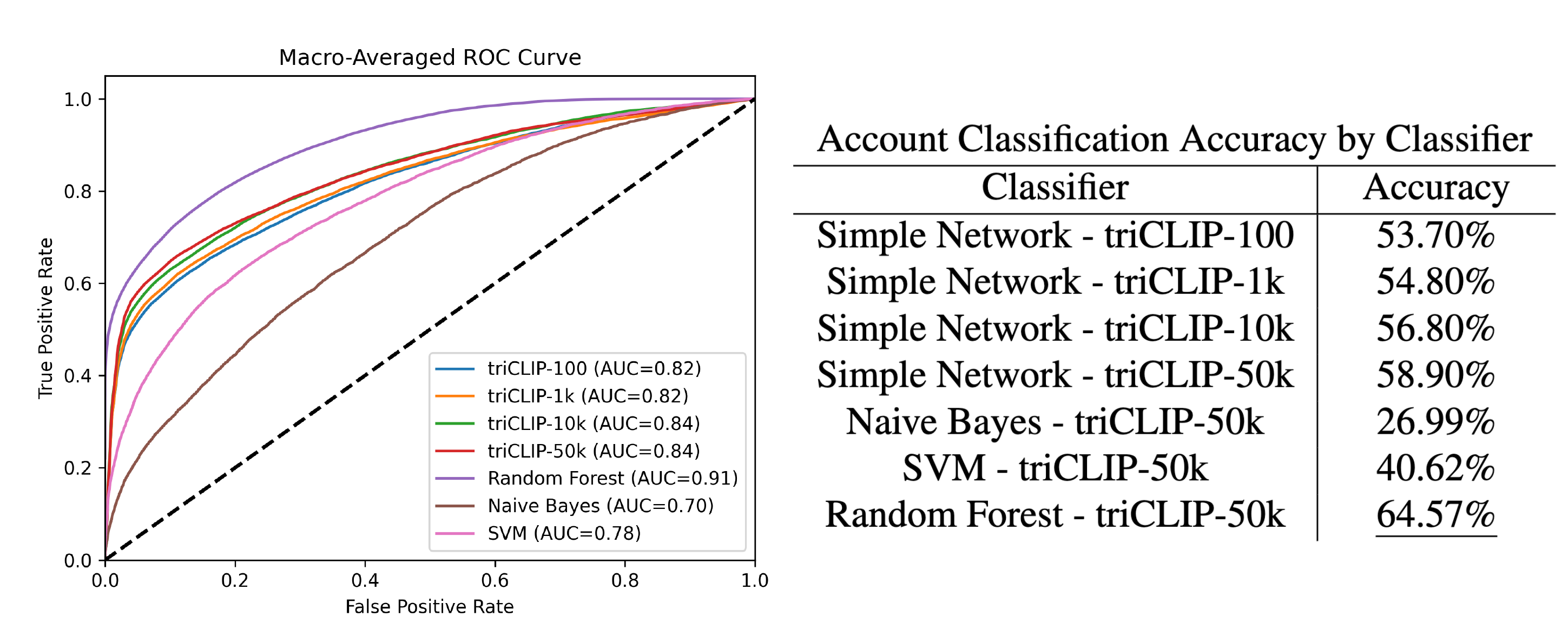} 
\caption{The ROC curves and AUC for the account classifiers, alongside the per-method classification accuracies. Random Forests achieved a 64.57\% accuracy across 10,000 posts when using triCLIP-50k features. Much like with the binary classifier, Naive Bayes performed significantly worse than any of the other methods.}
\vspace{-2mm}
\label{fig:acc_class}
\end{figure*}

\section{Applications in Trimodal Space}
\label{sec:apps}

To demonstrate the usefulness of N-dimensional CLIP loss applied to OSINT scenarios we demonstrate two applications of an N-dimensional model: a stance classifier for pro-Ukrainian or pro-Russian social media artifacts and a provenance classifier, labeling an artifact as belonging to a certain Telegram account.

\textbf{Binary Stance Classifier:} We demonstrate binary stance classifiers using our trimodal model to extract embeddings which allows the classification of videos, images, and text as either pro-Russian or pro-Ukrainian. Figure~\ref{fig:stance_class} gives the ROC curves for the classifiers using the various models trained. The models were evaluated using a 5-fold cross validation technique. The baseline accuracy of a binary classifier is 50\% and our classifiers achieve accuracies of 94.23\%. While we train four single-purpose binary classifiers following Tam's example~\cite{Tam_2023_binary}, the encodings from the triCLIP models could be utilized by any of the other common binary classification methods such as random forest~\cite{forests}, naive Bayes~\cite{bayes}, or SVMs~\cite{svm}. The four models trained were each given features from a different triCLIP model, with the amount of training data the model saw varying. This was to explore the effect that more training data had on a downstream classification task. Interestingly, using features from a model with more training data did not appear to improve the classification accuracy (the models used for feature extraction were all trained for 50 epochs).

Table~\ref{tab:binary_acc} shows the classification results of our model compared to the three other common methods. Our simple model matches the best AUC result by a baseline (Random Forest) with an AUC of 0.86. The results here are not meant to show that our simple biary classification model is state-of-the-art but instead meant to demonstrate the usefulness of features extracted using the triCLIP models developed above. With these new models, operators can classify the stance of an image, a text, or a video with a high degree of accuracy using simple out of the box binary classifiers with the features for all three modalities coming from a single model.

\textbf{Account Provenance Classifier:} We also show a multi-class account classifier using the triCLIP models, allowing the user to identify probabilistically what account a given artifact originated from. Figure~\ref{fig:acc_class} shows the accuracy on the train and test sets as a function of the number of epochs trained for. 

The final accuracy at epoch 5000 was 67.9\%. With 33 accounts in the testing set the baseline random accuracy is 3.03\% showing that the classifier works quite well. Table~\ref{tab:acc_class}  shows the classification results alongside three other methods of multi-class classification (KNN, Naive Bayes, and SVM). Due to the large imbalance between the number of posts each account has in the dataset, we make use of a synthetic minority over-sampling technique (SMOTE) by Chawla et al.~\cite{SMOTE} to artificially balance the dataset.


\begin{table}[]
\centering
\resizebox{\columnwidth}{!}{%
\begin{tabular}{ccccccccc}
\multicolumn{9}{c}{Effect of Projection Layer Dimensions on Recall @K for triCLIP-10k (5 Trials)} \\ \hline
\multicolumn{1}{c|}{Epochs} & \multicolumn{4}{c|}{10} & \multicolumn{4}{c}{50} \\ \hline
\multicolumn{1}{c|}{@K} & @1 & @5 & @10 & \multicolumn{1}{c|}{@25} & @1 & @5 & @10 & @25 \\ \hline
\multicolumn{1}{c|}{64D} & 24.60\% & 53.27\% & 65.87\% & \multicolumn{1}{c|}{76.80\%} & 58.67\% & 68.80\% & 75.20\% & 87.00\% \\
\multicolumn{1}{c|}{128D} & 22.40\% & 52.93\% & 63.60\% & \multicolumn{1}{c|}{76.07\%} & 56.53\% & 73.13\% & 78.33\% & 87.33\% \\
\multicolumn{1}{c|}{256D} & 24.53\% & 55.00\% & {\ul 68.60\%} & \multicolumn{1}{c|}{{\ul 83.13\%}} & {\ul 60.47\%} & {\ul 76.27\%} & {\ul 80.87\%} & {\ul 88.33\%} \\
\multicolumn{1}{c|}{512D} & {\ul 27.73\%} & 54.40\% & 65.27\% & \multicolumn{1}{c|}{79.53\%} & 54.43\% & 71.33\% & 77.00\% & 86.53\% \\
\multicolumn{1}{c|}{768D} & 27.20\% & {\ul 55.27\%} & 64.93\% & \multicolumn{1}{c|}{77.87\%} & 54.13\% & 71.07\% & 76.73\% & 85.00\%
\end{tabular}%
}
\caption{The effects of the size of the projection layer output on recall and train time. The model tested was trained on 10,000 posts at 10 and 50 epochs. Interestingly, a higher output dimensionality does not directly correlate to a higher recall.}
\label{tab:proj_recall}
\end{table}

\begin{table*}[t]
\resizebox{\textwidth}{!}{%
\centering
\begin{tabular}{ccccccccccccccccc}
\multicolumn{17}{c}{Average Recall for quadCLIP Models @K (5 Trials)} \\ \hline
\multicolumn{1}{c|}{Model} & \multicolumn{4}{c|}{quadCLIP-100} & \multicolumn{4}{c|}{quadCLIP-1k} & \multicolumn{4}{c||}{quadCLIP-10k} & \multicolumn{4}{c}{triCLIP-10k} \\ \hline
\multicolumn{1}{c|}{Epochs} & @1 & @5 & @10 & \multicolumn{1}{c|}{@25} & @1 & @5 & @10 & \multicolumn{1}{c|}{@25} & @1 & @5 & @10 & \multicolumn{1}{c||}{@25} & @1 & @5 & @10 & @25\\ \hline
\multicolumn{1}{c|}{1} & 0.67\% & 7.00\% & 13.33\% & \multicolumn{1}{c|}{30.00\%} & 0.67\% & 4.00\% & 9.00\% & \multicolumn{1}{c|}{25.67\%} & 2.00\% & 12.67\% & 19.67\% & \multicolumn{1}{c||}{37.33\%} & 4.33\% & 17.53\% & 27.13\% & 49.90\% \\
\multicolumn{1}{c|}{10} & 0.33\% & 3.00\% & 7.00\% & \multicolumn{1}{c|}{26.00\%} & 2.67\% & 14.00\% & 21.67\% & \multicolumn{1}{c|}{41.00\%} & 10.00\% & 27.33\% & 36.00\% & \multicolumn{1}{c||}{52.00\%} & 24.53\% & 55.00\% & 68.60\% & 83.13\% \\
\multicolumn{1}{c|}{50} & 2.67\% & 6.33\% & 14.67\% & \multicolumn{1}{c|}{34.67\%} & 5.00\% & 18.67\% & 29.67\% & \multicolumn{1}{c|}{48.33\%} & 15.67\% & 37.00\% & 47.00\% & \multicolumn{1}{c||}{61.67\%} & 60.47\% & 76.27\% & 80.87\% & 87.27\%
\end{tabular}
}
\caption{The recall results for the quadCLIP model (and triCLIP-10k for easy comparison). They are noticeably worse than the results for triCLIP models at the same number of epochs and training data but further exploration is required to determine the reason.}
\label{tab:quad_recall}
\end{table*}

\section{Quadmodal Contrastive Loss}
\label{sec:quad_poc}

To demonstrate the fact that the N-dimensional triplet loss formalized above extends beyond the trimodal applications shown above we train a simple proof-of-concept model on tuples with 4 modalities, the first of its kind to our knowledge. In addition to image, video, and text we add audio. As outlined above, current video encoders simply sample several frames from a video and treat them as images. Due to this process the audio of a video plays no role in the generation of the embedding and is more or less entirely cast aside. By extending our model to 4 dimensions, via adding an additional audio encoder that can take in the audio from a video, we can avoid losing the information that the signal provides and use it as another positive sample in our tuple.

We use an audio encoder similar to the ones used for the other three modalities: 'wav2vec2'~\cite{wav2clip}. Using this encoder we can extract an additional audio embedding, which is ignored when the video embedding is generated. The audio is sampled at 16000 Hz. and then from the audio track a single 768 dimensional vector is extracted. Using a variation of Equation~\ref{eq:nclip} pegged to four modalities we can train a model following the same process as done for the triCLIP models.

Table~\ref{tab:quad_recall} gives the retrieval results for the quadCLIP models, evaluated in the same manner as the triCLIP models from Section~\ref{sec:method}. As the model is intended more as a proof-of-concept we train on sets of only 100, 1000, and 10,000 items and for only 1, 10, and 50 epochs. The best results were achieved by the model trained for the most epochs on the most data. Similarly to triTRIP models, the results at K=1 is much lower than its triCLIP counterpart, though they get closer as K grows larger. While these results demonstrate that as is, the quadCLIP model could prove useful, more exploration is needed to understand why quadCLIP experiences the drop in retrieval, especially at lower values of K.

While theoretically the N-dimensional contrastive loss could be scaled to an infinite number of modalities, it seems likely a ceiling would quickly take effect. In reality there are only so many modalities that are present on social media, so until scratch-and-sniff monitors are released, scaling above four dimensions seems unlikely. Our novel results provide a benchmark for future efforts in the quad-modal space.

\section{Future Work and Conclusions}
\label{sec:concs}

\textbf{Future Work:} We demonstrate that trimodal contrastive loss models can successfully be applied to social media. A possible avenue for future work is exploring the effects the encoders have on their individual modalities, in addition to jointly training encoders on the data. This work leveraged out of the box encoders and performed no pre-training. One could expect the results to improve were this done.

The results of quadCLIP on retrievaly show a drop relative to similar trimodal models. The underlying causes of this are unclear, though it would be natural to suspect it's something to do with the audio embeddings. More exploration needs to be done on successfully embedding audio for use in a quadmodal model. New datasets of social media data where each post contains more than two modalities would also be of use to future efforts in this space.

\textbf{Conclusions:} To further our computational analysis of social media it's of vital importance to take the entire context of a post into consideration, which includes studying all of its modalities. The amount of videos available on social media only continues to grow. By extending contrastive loss to N-dimensions we can theoretically enable the comparison of any number of modalities on social media. High accuracies on media retrieval is useful for OSINT operators trying to understand and discover similar posts in fast-moving multi-modal situations.

In addition to an extension of contrastive loss and a display of its usefulness in retrieval settings, we demonstrate the usefulness of the trimodal extension to various use-cases on social media. Using a triCLIP model allows the classification of three different modalities using a singular model. We show results on two different classification tasks: a binary stance classification task (pro-Ukrainian or pro-Russian) and a multiclass classification task of account provenance.

We also demonstrate, a first to our knowledge, the extension of CLIP loss to four modalities. The results of quadCLIP provide a new baseline for others in this space moving forward. Having a single model that can handle any modality seen on social media and measure similarity across the modalities is valuable to many non-technical workers in the OSINT space and extending the tooling on top of triCLIP and quadCLIP models presented here is an exciting task.


{\small
\bibliographystyle{ieee_fullname}
\bibliography{egbib}
}

\end{document}